\documentclass{article}

\usepackage{arxiv}

\usepackage{url}            
\usepackage{booktabs}       
\usepackage{amsfonts}       
\usepackage{nicefrac}       
\usepackage{microtype}      

 \usepackage{hyperref}
 \hypersetup{colorlinks=true,linkcolor=citeblue,citecolor=citeblue,urlcolor=citeblue}

\usepackage{cite}
\usepackage{amsmath,amssymb,amsfonts}
\usepackage{cleveref}
\usepackage{subcaption}
\usepackage{makecell}

\usepackage[T1]{fontenc}
\usepackage{lmodern}

\usepackage{lipsum}
\usepackage{xcolor}
\usepackage{graphicx}
\usepackage{geometry}
\usepackage{array}
\usepackage{multirow}
\usepackage{xcolor}
\definecolor{citeblue}{RGB}{0,0,255}

\graphicspath{ {./images/} }

\title{Unraveling the Dominance of Large Language Models Over Transformer Models  for Bangla Natural Language Inference: A Comprehensive Study}

\author{
\\
\newline
\\
Fatema Tuj Johora Faria \textsuperscript{*},
Mukaffi Bin Moin,
Asif Iftekher Fahim,
Pronay Debnath,
\\
Faisal Muhammad Shah
\\
\bigskip
\\
\\Ahsanullah University of Science and Technology, Dhaka, Bangladesh.
\\
\\
\\
\bigskip
*Corresponding author(s). E-mail(s): \texttt{\textcolor{blue}{fatema.faria142@gmail.com}}\\
Contributing authors: \texttt{\textcolor{blue}{mukaffi28@gmail.com}}; \texttt{\textcolor{blue}{fahimthescientist@gmail.com}}; \\\texttt{\textcolor{blue}{pronaydebnath99@gmail.com}}; \texttt{\textcolor{blue}{faisal.cse@aust.edu}}; 
\\ 
}

\begin{document}

\maketitle
\abstract{
Natural Language Inference (NLI) is a cornerstone of Natural Language Processing (NLP), providing insights into the entailment relationships between text pairings. It is a critical component of Natural Language Understanding (NLU), demonstrating the ability to extract information from spoken or written interactions. NLI is mainly concerned with determining the entailment relationship between two statements, known as the premise and hypothesis. When the premise logically implies the hypothesis, the pair is labeled ``entailment''. If the hypothesis contradicts the premise, the pair receives the ``contradiction'' label. When there is insufficient evidence to establish a connection, the pair is described as ``neutral''. Despite the success of Large Language Models (LLMs) in various tasks, their effectiveness in NLI remains constrained by issues like low-resource domain accuracy, model overconfidence, and difficulty in capturing human judgment disagreements. This study addresses the underexplored area of evaluating LLMs in low-resourced languages such as Bengali. Through a comprehensive evaluation, we assess the performance of prominent LLMs and state-of-the-art (SOTA) models in Bengali NLP tasks, focusing on natural language inference. Utilizing the XNLI dataset, we conduct zero-shot and few-shot evaluations, comparing LLMs like GPT-3.5 Turbo and Gemini 1.5 Pro with models such as BanglaBERT, Bangla BERT Base, DistilBERT, mBERT, and sahajBERT. Our findings reveal that while LLMs can achieve comparable or superior performance to fine-tuned SOTA models in few-shot scenarios, further research is necessary to enhance our understanding of LLMs in languages with modest resources like Bengali. This study underscores the importance of continued efforts in exploring LLM capabilities across diverse linguistic contexts.Our implementations are now available for public access at: \href{https://github.com/fatemafaria142/Large-Language-Models-Over-Transformer-Models-for-Bangla-NLI}{{https://github.com/fatemafaria142/Large-Language-Models-Over-Transformer-Models-for-Bangla-NLI}}}



\section{Introduction}\label{introlab}
During human communication, a lot of information is conveyed. Typically, the receiver learns far more than what the speaker says. This is due to humans' innate inference ability or broad world knowledge. Understanding the relationship between premise and hypothesis is important in NLI. Discordances between them might describe the implication relationship as contradictory \cite{Intro1}.

Recent advances in LLMs \cite{Intro2} \cite{intronew3} have ushered in a new era of NLP, bringing unparalleled possibilities. These LLMs, with parameter sizes reaching a hundred billion, use contextual learning to thrive in few-shot and zero-shot learning scenarios, removing the need for task-specific fine-tuning. This innovation not only eliminates dependency on labeled datasets, but also allows the model to perform magnificently on a wide range of tasks for which it was not expressly trained. Despite their impressive capabilities, LLMs are prone to producing erroneous data, necessitating the use of techniques such as Reinforcement Learning from Human Feedback (RLHF) \cite{Intro3} to assure the development of dependable responses.

Autoregressive language models \cite{Intro4}, epitomized by the Generative Pre-trained Transformer series like GPT-3, have revolutionized NLP approaches. Operating as a decoder-only transformer, GPT-3 excels at generating text sequentially, leveraging attention mechanisms to focus on pertinent segments of input text. This model demonstrates exceptional versatility, adeptly tackling various language tasks without task-specific fine-tuning and excelling in zero-shot and few-shot learning scenarios. GPT-3's capability enables it to respond effectively to text-based prompts, even with limited examples or instructions, facilitating a more interactive and dynamic text generation process.

ChatGPT \cite{GPT} and another recently proposed LLM, like Gemini 1.5 Pro \cite{Gemini}, have garnered significant attention for their utilization of the RLHF technique to overcome various limitations observed in earlier models. While these models are trained across multiple languages, English remains dominant in their training datasets. Despite demonstrating robust zero-shot performance across various NLP tasks in English and other languages, their application in the Bengali language domain, which is widely spoken yet resource-constrained, remains largely unexplored. The use of LLMs in prevalent downstream NLP tasks specific to Bengali has not been thoroughly investigated compared to the extensive evaluations conducted for English and other Indo-European languages. Due to the shortage of annotated datasets tailored to Bengali, we conducted a zero-shot and few-shot evaluation with LLMs to assess their performance. The findings presented in this paper offer valuable insights into the capabilities and limitations of LLMs in Bengali for NLI. 
Our study reveals the following key findings:
\begin{itemize}
\item 
We observe that LLMs generally demonstrate lower zero-shot performance compared to state-of-the-art (SOTA) fine-tuned models, especially in the context of Bangla NLI. Despite occasional outliers, this performance gap suggests that existing LLMs may not adequately meet the requirements of low-resource Bengali tasks, particularly in Natural Language Inference (NLI). Additionally, the phenomenon of ``hallucination,'' where LLMs generate incorrect or inappropriate information presented as factual, poses an additional challenge in ensuring the reliability of their outputs.

\item We find it noteworthy that LLMs demonstrate impressive performance in zero-shot scenarios within the English language. However, their performance in languages with fewer resources, such as Bengali, falls short. This underscores the significance of delving into the constraints faced by LLMs customized for various modest-resource language communities.

\item Furthermore, our findings reveal that the performance of LLMs significantly enhances with just a few shots (5 shot, 10 shot, 15 shot) compared to zero-shot scenarios. Surprisingly, this improvement surpasses even the performance of SOTA models. This underscores the vast potential of LLMs and indicates promising avenues for further exploration and enhancement.

\end{itemize}

\section{Related Works} \label{Relatedlab}
\label{sec:2}
Mohsinul et al. \cite{Intro2} evaluate LLMs in Bengali using BenLLM-Eval, assessing GPT-3.5, LLaMA-2-13b-chat, and Claude-2 across various NLP tasks. Results show mixed performance, with some tasks matching or surpassing SOTA, notably in sentiment analysis, while others, especially with LLaMA-2-13b-chat, exhibit poor performance. In text summarization, GPT-3.5 produces longer summaries, contrasting Claude-2's concise outputs. LLaMA-2-13b-chat often generates English summaries, impacting its scores significantly. On the other hand, Reto et al. \cite{Related2} discovered that LLMs, when fine-tuned on MNLI, struggled to recognize the inability of questions, commands, or incomprehensible fragments to infer or contradict information. However, their experiments with new fine-tuning datasets showed promising results. By specifically addressing this issue, LLMs demonstrated improved sensitivity to non-assertive premises without significantly sacrificing accuracy on MNLI-matched tasks. This suggests a potential avenue for enhancing LLMs' understanding of logical inference from diverse prompts. Additionally, Xuanli et al. \cite{Related3} introduced the X-ICL (ChatGPTfew) method, leveraging human-written natural language explanations (NLEs) to prompt ChatGPT to generate synthetic NLEs. They found that this approach significantly improves accuracy across various adversarial NLI datasets and five LLMs compared to standard in-context learning and X-ICL using human-written NLEs. Furthermore, their analysis highlighted potential limitations of data selection methodologies in handling adversarial datasets, emphasizing the consistent performance of their approach in both in-distribution and adversarial contexts.  Lastly, Noah et al. \cite{Related4} assessed LLMs' performance and alignment with human disagreement in NLI using MCE and LPE methods. They discovered LLMs struggle in NLI tasks and fail to capture human disagreement, especially in cases with high disagreement levels. GPT/Flan-T5-XXL showed the best NLI capability and alignment with human disagreement when using LPE/MCE. Their contributions include pioneering joint evaluation of LLMs' performance and human disagreement alignment, proposing estimation techniques, and exploring model sensitivity to estimation methods and prompts, utilizing JSD as a key metric.

\section{Background Study} \label{Backgroundlab}
\subsection{Dataset Collection}
In this research paper, we have employed the ``XNLI'' \cite{Dataset} dataset, which comprises Bengali-language instances for NLI. The dataset consists of a subset of the MultiNLI data, where each instance includes a premise (sentence1), a hypothesis (sentence2), and a classification label indicating contradiction (0), entailment (1), or neutral (2). We utilized 381,449 instances from the train set, 2,419 from the validation set, and 4,895 from the test set. 

\subsection{Models}
\subsubsection{Pre-trained Language Models (PLMs) on Bangla}
Developing language-specific models poses significant challenges, particularly for languages with limited resources like Bengali. Nonetheless, several pretrained language models have recently emerged, gaining recognition for their SOTA performance across diverse downstream tasks. The discussion below briefly outlines their achievements, while Table \ref{tab:pretrained_lm} presents the configurations of Bengali pretrained language models. \\
\textbf{1) ELECTRA based:}
ELECTRA, a powerful pre-trained language model, employs a unique method called ``Replaced Token Detection (RTD)'' instead of traditional self-attention mechanisms. It achieves computational efficiency by generating representations solely for replaced tokens, not the entire input sequence. BanglaBERT \cite{Dataset} generator, a variant of ELECTRA, is pre-trained using the masked language modeling objective on extensive Bengali corpora.\\
\textbf{2) BERT based:}
BERT, short for Bidirectional Encoder Representations from Transformers, is a pre-trained language model that revolutionized NLP. It introduced a bidirectional approach to context understanding, capturing deeper semantic meaning by considering both left and right context simultaneously. Bangla BERT Base \cite{banglabert}, DistilBERT \cite{Distil}, and mBERT \cite{mbert} are all pre-trained Bengali language models built on the mask language modeling framework, as pioneered by BERT.\\
\textbf{3) ALBERT based:}
ALBERT prioritizes scalability by minimizing model size and computational requirements while improving performance. Through techniques like parameter reduction and sharing, it surpasses traditional BERT models in efficiency. Its Lite architecture, with parameter sharing and factorized embedding, ensures quality with fewer parameters. SahajBERT\footnote{\href{https://huggingface.co/neuropark/sahajBERT}{{https://huggingface.co/neuropark/sahajBERT}}}, a collaborative effort for Bengali, is an ALBERT model pre-trained using masked language modeling methods.

\begin{table}[ht]
    \centering
    \caption{Configurations of Bengali Pretrained Language Models}
    \vspace{10pt}
    \label{tab:pretrained_lm}
    \begin{tabular}{p{3cm} p{3cm} p{2.8cm} p{2.0cm} p{2.5cm}} 
        \toprule
        \textbf{Model} & \textbf{Objective During \newline Training} & \textbf{Type} & \textbf{Parameters} & \textbf{Embedding Size} \\ \midrule
        BanglaBERT & MLM with RTD & ELECTRA-based & 110M & 768 \\
        Bangla BERT Base & MLM & BERT-based & 110M & 768 \\
        DistilBERT & MLM & BERT-based & 66M & 768 \\
        mBERT & Multilingual MLM & BERT-base & 110M & 768 \\
        sahajBERT & MLM & ALBERT-based & 18M & 128 \\ \bottomrule
    \end{tabular}
\end{table}

\subsubsection{Large Language Models (LLMs) on Bangla}
\textbf{1) GPT-3.5 Turbo:}
 {GPT-3.5 Turbo \cite{GPT}},  Turbo, the flagship model in the GPT-3.5 series, powers the free version of ChatGPT, providing a cost-effective and flexible solution for text generation tasks. With a context window of 16,385 tokens and improved accuracy in requested formats, it addresses text encoding issues for non-English functions and excels in delivering high-quality responses within an efficiency limit of 4,096 output tokens. \\
\textbf{2) Gemini 1.5 Pro:}
Gemini 1.5 Pro \cite{Gemini}, from Google DeepMind's Gemini family, is a cutting-edge multi-modal model adept at processing text, audio, and video, enhancing its versatility across tasks. Its standout feature is its proficiency in long-context understanding, enabled by a 128,000 token context window, facilitating nuanced comprehension and insightful responses. 

\subsection{Evaluation Metrics}
In NLI, accuracy refers to the overall correctness of predictions across all classes (neutral, entailment, and contradiction) as measured by the proportion of properly categorized examples out of the total. Precision indicates the model's precision when predicting a certain class, representing the proportion of true positive predictions for that class among all occurrences anticipated to be that class. Recall measures the model's capacity to accurately identify examples of a certain class by calculating the proportion of true positive predictions for that class among all instances that belong to it. The F1 Score gives a fair assessment of the model's performance across several classes, taking into account both accuracy and recall, making it very valuable for dealing with unbalanced classes in NLI tasks.

\section{Implementation Details} \label{Methodologylab}
\subsection{Experiment 1: Natural Language Inference using PLMs}

\textbf{Step 1) Text Preprocessing:} We implement a series of preprocessing steps, including normalization, on the Bangla XNLI dataset to ensure compatibility with PLM input requirements. In this process, we utilize a text normalization module specifically designed for cleaning Bengali text. Our normalization procedure includes handling multiple white spaces, commas, and unicode characters, as well as correcting single or double quotes throughout the dataset.\\
\textbf{Step 2) Model Initialization:} Model Initialization entails the setup of pre-trained Bengali language models along with their respective weights and configurations.\\
\textbf{Step 3) Fine-tuning Procedure:} During the Fine-tuning Procedure, we take the initialized PLMs and subject them to training on the Bangla XNLI dataset using transfer learning techniques. This process involves updating the model parameters through gradient descent optimization algorithms, with the objective of minimizing a suitable loss function tailored specifically to NLI tasks. In particular, we define the optimizer as AdamW and the loss function as CrossEntropyLoss.\\
\textbf{Step 4) Training Settings:} Hyperparameters, essential settings in deep learning models, shape the learning process and impact performance and generalization. During fine-tuning, adjustments to parameters like learning rate, batch size, and epochs are made to optimize performance and mitigate overfitting. Learning rate governs optimization step size, batch size dictates sample processing per iteration, and epochs determine dataset utilization. Careful tuning of these hyperparameters ensures effective training while guarding against excessive memorization. Table \ref{hype} offers detailed guidance for hyperparameter optimization, enhancing model performance.\\
\textbf{Step 5) Performance Assessment:} We evaluate the fine-tuned PLMs on a held-out test set using defined evaluation criteria, resulting in an objective assessment of their performance on Bangla NLI tasks. Metrics like Accuracy, Precision, Recall, and F1 score are used to correctly assess model performance. Table \ref{table1} shows comprehensive information, including these measures, that provide light on the performance of the fine-tuned models for Bangla NLI tasks.

\subsection{Experiment 2: Natural Language Inference using LLMs}
\subsubsection{Data Selection}
We randomly selected 300 data points, extracted from the training set of the XNLI dataset, to evaluate the performance of Zero Shot and Few Shot prompts. These data points were divided into 100 instances each for neutral, entailment, and contradiction. To mitigate costs associated with using the OpenAI GPT-3.5 Turbo API and Gemini 1.5 Pro API, we conducted our experiments on this subset instead of the entire collection. Additionally, we only considered data points where both the premise and hypothesis contained more than or equal to 3 words.
\subsubsection{Prompting Template}
\begin{figure}
    \centering
    \includegraphics[width=1\textwidth, height=350px]{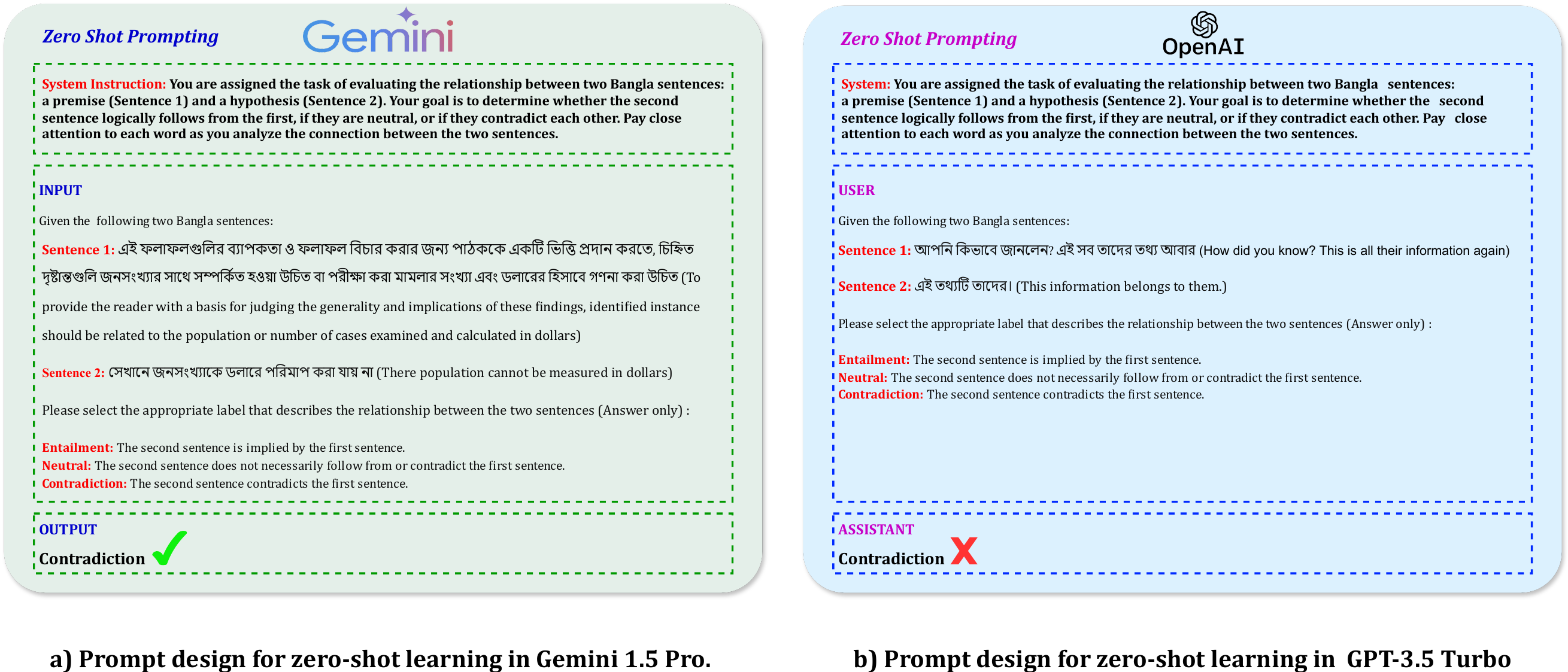}
    \caption{Illustration of innovative zero-shot prompt designs for Gemini 1.5 Pro and GPT-3.5 Turbo exemplify effective methodologies in Bangla NLI.}
    \label{fig:dia}
\end{figure}

\textbf{1) Zero-shot prompting for Bangla NLI:}
Zero-shot prompting in NLI involves constructing prompts to guide the model in predicting the logical relationship between a given premise and hypothesis, without explicit training on specific examples of this relationship. The premise offers context, while the hypothesis presents a claim based on that context. The model aims to determine whether the hypothesis logically follows (entailment), contradicts (contradiction), or has no strong logical connection (neutral) to the premise. This process includes prompt construction, premise-hypothesis pairing, and prediction. Figure \ref{fig:dia} elegantly displays both GPT-3.5 Turbo's Zero Shot prompting and Gemini Pro's exemplary performance in Bangla NLI. It effectively categorizes premise-hypothesis pairs, showcasing the accuracy of predictions in Bangla NLI without the need for explicit training.\\
\textbf{2) Few-shot prompting for Bangla NLI:}
Few-shot prompting in NLI involves providing the model with a limited number of example premise-hypothesis pairs to guide its predictions regarding the logical relationship between subsequent pairs, categorized as contradiction, entailment, or neutral. The premise offers context, while the hypothesis presents a claim based on that context. The model is then tasked with determining the appropriate relationship between them. In a 5-shot scenario, the model receives five such pairs for training before making predictions on new data, while in a 10-shot scenario, it receives ten pairs, and so on.

\section{Result Analysis}\label{resultlab}
\subsection{Hyperparameter Settings}
The Table \ref{hype} outlines hyperparameter configurations for NLI with the XNLI dataset, encompassing BanglaBERT, Bangla BERT Base, DistilBERT, mBERT, and sahajBERT. We experimented with common ranges for learning rates (0.001 to 0.01), batch sizes (8 to 16), and epochs (5 to 20), but settled on a standardized setup of a learning rate of 0.001, batch size of 16, and training for 15 epochs. Each model undergoes training using the Adam optimizer, ensuring consistency across experiments and facilitating fair performance comparison.
\begin{table}[ht]
\caption{Experimenting with Hyperparameters on ``XNLI'' Dataset}
\vspace{10pt}
\label{hype}
\centering
\begin{tabular}{lllll} 
\toprule
\textbf{Models} & \textbf{Learning Rate} & \textbf{Batch Size} & \textbf{Number of Epochs} & \textbf{Optimizer} \\
\midrule
BanglaBERT & \centering  0.001 & \centering  16 & \centering 15 &  Adam \\
Bangla BERT Base & \centering  0.001 & \centering 16 & \centering 15 & Adam \\
DistilBERT & \centering  0.001 & \centering 16 & \centering 15 & Adam \\
mBERT & \centering  0.001 & \centering 16 & \centering 15 & Adam \\
sahajBERT & \centering  0.001 & \centering 16 & \centering 15 & Adam \\
\bottomrule
\end{tabular}
\end{table}

\label{sec:5}
\subsection{Performance Analysis of PLMs}
In the domain of Bangla NLI, BanglaBERT emerges as the unequivocal leader, as demonstrated in Table \ref{table1}. With an impressive accuracy of 82.04\%, BanglaBERT not only achieves the highest accuracy among all models but also showcases superior precision, recall, and F1-Score metrics. This robust performance underscores its remarkable ability to capture the nuances present in Bangla text. On the contrary, DistilBERT lags behind with the lowest accuracy recorded at 63.20\%, while mBERT exhibits the least favorable F1-Score at 61.53\%. Nevertheless, it's essential to highlight that despite these variations, all models maintain precision and recall metrics above 63\% and 63.20\%, respectively, indicating a consistent level of performance across the models.
\begin{table}[h]
\caption{Comparative Analysis of PLMs for Different Performance Metrics}
\label{table1}
\centering
\begin{tabular}{lllll} 
\toprule
\textbf{Model} & \textbf{Accuracy} & \textbf{Precision} & \textbf{Recall} & \textbf{F1-Score} \\
\midrule
\textbf{BanglaBERT} & \textbf{0.8204} &  \textbf{0.8222} &  \textbf{0.8204} &  \textbf{0.8203} \\
Bangla BERT Base & 0.6803 & 0.6907 & 0.6812 & 0.6833 \\
DistilBERT & 0.6320 & 0.6358 & 0.6320 & 0.6317\\
mBERT & 0.6427 & 0.6496 & 0.6428 & 0.6153\\
sahajBERT & 0.6708 & 0.6791 & 0.6709 & 0.6707\\
\bottomrule
\end{tabular}
\end{table}

\subsection{Case Study Take-Away of LLMs}

\textbf{What elements should be included in a well-crafted prompt?}
Crafting a precise Bangla NLI prompt necessitates clear instructions, incorporating both premise and hypothesis for inference. Attention to language components and their relationships is crucial for accurate conclusions. Strategic use of keywords directs the model's focus, enhancing its inference capabilities.\\
\textbf{What are the potential reasons behind the phenomenon of zero-shot collapse?}
The collapse of zero-shot prompting in Bangla NLI arises from limited pre-training data and dataset biases, hampering generalization. Furthermore, the intricate nature of NLI, compounded by Bangla language nuances, poses significant hurdles for the model. Capturing these language specifics effectively proves challenging, amplifying the model's struggle in forming accurate inferences and leading to the collapse phenomenon.\\
\textbf{Why is few-shot learning considered advantageous?}
Few-shot learning in Bangla NLI offers adaptability and generalization with minimal supervision, crucial where labeled data is scarce. In 5-shot learning, five examples per task aid in grasping linguistic patterns, reducing dependency on large datasets. 10-shot learning expands this capability, enhancing task-specific understanding within the Bangla context. With 15-shot learning, increased examples bolster robust adaptation and generalization across diverse tasks.\\
\textbf{How is the effectiveness of a prompt evaluated?}
We evaluate Bangla NLI prompts based on their ability to guide models in inferring logical relationships between premise and hypothesis. Performance metrics like accuracy, precision, recall, and F1 score assess prompt efficacy on test sets. Our control settings for Large Language Models include Temperature (1.0), Top P (1.0), Maximum tokens (256), Frequency penalty (0.0), Presence penalty (0.0). We summarize Zero-shot and Few-shot learning results in Table \ref{table10}, detailing performance metrics across inference tasks, providing insights into the effectiveness of these strategies in NLI. \\
\textbf{What are the current limitations regarding hallucinations in Bangla text for Language Models?} Hallucinations in LLMs refer to instances where the model generates responses that are factually incorrect, nonsensical, or disconnected from the input prompt. These occurrences stem from the probabilistic nature of LLMs, which base responses on learned patterns rather than true understanding. While we've observed such inaccuracies in Bangla text, insights from other languages suggest their relevance. The limited training data for Bangla compared to more well-resourced languages like English may contribute to these hallucinations, impeding nuanced language comprehension. Interestingly, the frequency of hallucinations appears to be higher in Gemini 1.5 Pro compared to GPT 3.5 Turbo.
\begin{table}[ht]
\caption{Comparative Analysis of LLMs for Different Performance Metrics}
\label{table10}
\centering
\begin{tabular}{llllll} 
\toprule
\textbf{LLMs} & \textbf{Metric} & \textbf{Zero-shot} & \textbf{5-shot} & \textbf{10-shot} & \textbf{15-shot} \\
\midrule
\multirow{4}{*}{GPT-3.5 Turbo} & Accuracy & 0.7403 & 0.8657 & 0.8756 & \textbf{0.9205} \\
& Precision & 0.6825 & 0.8683 & 0.8753 & \textbf{0.9219} \\
& Recall & 0.7234 & 0.8624 & 0.8759 &  \textbf{0.9204} \\
& F1-Score & 0.8054 & 0.8640 & 0.8748 & \textbf{0.9299} \\
\midrule
\multirow{4}{*}{Gemini 1.5 Pro} & Accuracy & 0.7287 & 0.8625 & 0.8732 & \textbf{0.9146} \\
& Precision & 0.5556 & 0.8652 & 0.8763 & \textbf{0.9156} \\
& Recall & 0.7143 &  0.8652 & 0.8732 & \textbf{0.9146} \\
& F1-Score & 0.6251 & 0.8652 & 0.8701 & \textbf{0.9136} \\
\bottomrule
\end{tabular}
\end{table} 

\section{Limitations and Future Research Directions}
\label{sec:6}
In our future work, we will address several notable limitations that may impact the fidelity and reliability of NLI statements generated by LLMs like GPT-3.5 and Gemini 1.5 Pro. While these models excel in providing accurate answers, we acknowledge instances where the generated NLI statements contain inaccuracies or fabricated information, which could undermine users' trust in the model. Moreover, the undisclosed datasets used for training these LLMs introduce uncertainty regarding the origins of evaluation datasets. Additionally, we recognize that the evaluation process of the GPT-3.5 model relies on OpenAI's GPT-3.5 turbo API, which, while cost-effective for token counting, may pose certain limitations. We understand that prompt engineering is crucial for optimizing the performance of LLMs. However, the manual design and refinement of prompts for NLI tasks are acknowledged as time-consuming endeavors. Although Chain-of-Thought (CoT) Prompting holds promise in addressing these limitations by guiding LLMs through multi-step reasoning processes, it was not incorporated into this study, resulting in a lack of insight into the model's decision-making process. Despite our meticulous efforts to optimize prompts, we acknowledge the challenge of navigating the vast prompt space to find the most effective prompts. We understand that suboptimal prompts may hinder the true capabilities of LLMs. Our future research aims to overcome this challenge by leveraging automatic prompt optimization algorithms to tailor task-specific and model-specific prompts, thereby enhancing LLM performance. While an improved version, GPT-4, exists, its reduced cost-effectiveness compared to GPT-3.5 makes it impractical for this study. Nevertheless, despite these limitations, our research lays the groundwork for future investigations into LLM performance in languages with modest resources within the NLP domain. We believe that future endeavors could benefit from integrating CoT Prompting into Bangla NLI tasks, offering valuable insights into enhancing reasoning abilities and addressing fidelity and transparency issues in NLI statements.

\section{Conclusion}
\label{sec:7}
In this research, we investigated Bangla Natural Language Inference (NLI) by conducting a comparative analysis of PLMs and LLMs. Our major goal was to thoroughly assess the performance of these models using important measures such as accuracy, precision, recall, and F1 score. We aimed to determine their efficacy through extensive testing. Our studies yielded surprising outcomes. Notably, GPT-3.5 Turbo achieved the greatest accuracy of 0.9205, followed by Gemini 1.5 Pro at 0.9146, which outperformed BanglaBERT's 0.8204. While PLMs showed promise, they eventually fell short when compared to LLMs, highlighting architectural complexities and fine-tuning issues. The disparity emphasizes the importance of continuing research and improvement of LLMs' competence in solving NLI tasks in Bengali. PLMs demonstrated proficiency, but did not surpass their larger equivalence This remark emphasizes the subtle interplay between model architecture, fine-tuning procedures, and the complex details of the task at hand. Such findings add not only to our understanding of LLM efficacy in Bengali NLI tasks, but also highlight the continuous need for research and development in this field.

\bibliographystyle{unsrt}
\bibliography{Mukaffi}

\end{document}